\RequirePackage{fix-cm}
\documentclass[smallextended]{svjour3}       
\smartqed  
\usepackage{graphicx}
\usepackage{amsmath}
\journalname{International Journal on Interactive Design and Manufacturing}
\begin{document}

\title{Human Arm simulation for interactive constrained environment design}


\author{
Liang MA \and
        Ruina MA \and
        Damien CHABLAT \and
        Fouad BENNIS
}


\institute{
Liang MA \at
              Department of Industrial Engineering, Tsinghua University, 100084, Beijing, China \\
              Tel.: +86-10-62792665 \\
              Fax: +86-10-62794399\\
              \email{liangma@tsinghua.edu.cn}           
             \emph{Present address:} of F. Author  
           \and
           Ruina MA, Damien CHABLAT, Fouad BENNIS \at
           Institut de Rechecherche en Communications et Cybernétique de Nantes, UMR CNRS 6597, 1 rue de la 
           \email{\{ruina.ma, damien.chablat, fouad.bennis\}@irccyn.ec-nantes.fr}
}

\date{Received: \today / Accepted: date}

\maketitle

\begin{abstract}
During the conceptual and prototype design stage of an industrial product, it is crucial to take assembly/disassembly and maintenance operations in advance. A well-designed system should enable relatively easy access of operating manipulators in the constrained environment and reduce musculoskeletal disorder risks for those manual handling operations. Trajectory planning comes up as an important issue for those assembly and maintenance operations under a constrained environment, since it determines the accessibility and the other ergonomics issues, such as muscle effort and its related fatigue. In this paper, a customer-oriented interactive approach is proposed to partially solve ergonomic related issues encountered during the design stage under a constrained system for the operator's convenience. Based on a single objective optimization method, trajectory planning for different operators could be generated automatically. Meanwhile, a motion capture based method assists the operator to guide the trajectory planning interactively when either a local minimum is encountered within the single objective optimization or the operator prefers guiding the virtual human manually. Besides that, a physical engine is integrated into this approach to provide physically realistic simulation in real time manner, so that collision free path and related dynamic information could be computed to determine further muscle fatigue and accessibility of a product design.  

\end{abstract}
\section{Introduction}
\label{Sec:1}
For industrial products, a compact design decreases the required massive space and enhances the appearance of the product. From another aspect, the designers have to consider constrained environments resulting from the compact design. Under constrained situations, assembly/disassembly oriented design has to be taken into consideration, since there are several ergonomic issues for the end user of the product or for the maintenance process. For example, the visibility \cite{SM} and accessibility \cite{LD,RS} of a component during an assembly operation; awkward posture caused by the product layout; physical or mental fatigue from the operations, etc.

For these reasons, virtual human simulations are often engaged during the conceptual design stage to evaluate the accessibility of the virtual environment and other ergonomic aspects \cite{MC1,RM}. Thanks to the interaction between the virtual human and the constrained environment, the designers are able to evaluate the manual handling operations, plan the possible trajectories, and further improve the design.

Trajectory planning \cite{LSM} is one of the most important problems for the use of virtual human in product design. In general, three approaches have been used frequently in the literature to generate the trajectory: inverse kinematics \cite{CP2009}, optimization-based method \cite{DPG} , and motion capture method \cite{MGC}. Inverse kinematics can generate a trajectory automatically and rapidly; however, this method could not generate a collision-free path easily. In order to overcome this inconvenience, an optimization-based approach has been proposed in \cite{RM} to find a collision-free path iteratively. In comparison to inverse kinematics, direct kinematics has been used in the optimization based approach and it enhances the computation efficiency. However, sometimes, the path can be trapped in a local minimum and it cannot get out from it without external intervention. Using haptic device or motion capture method, it is convenient to achieve natural movement in a virtual environment \cite{LR}. However, the motion data obtained from the motion capture has to be processed using motion retargeting method to adapt it to the overall population. 

In our research, we are aiming at creating a trajectory planning and evaluation method to improve the product design during the conceptual design stage. Virtual human modeling is taken to represent the overall population with different anthropometrical data. Ergonomic criterions such as fatigue is considered in the multi-objective evaluation. A single objective optimization based method is used to generate the trajectory at first. Then, motion capture methods or other intervention methods are used to help the algorithm to move out from the local minimum. At last, multi-objective evaluation methods are going to be used to evaluate the generated trajectory.
\section{Trajectory planning algorithm}
\label{sec:2}
A virtual human is modeled using the modified Denavit-Hartenberg method \cite{KD} with 28 degrees of mobilities to describe the mobility of all the key joints around human body. The kinematic information could be described by a set of general coordinates, where $q$ is the set of the rotational angles representing the positions of each joint. 
\subsection{- Single objective optimization}
\label{Sec:2_1}
\subsubsection{Trajectory planning algorithm}
\label{Sec:2_1_1}
\begin{figure*}
\centering
  \includegraphics[scale=1]{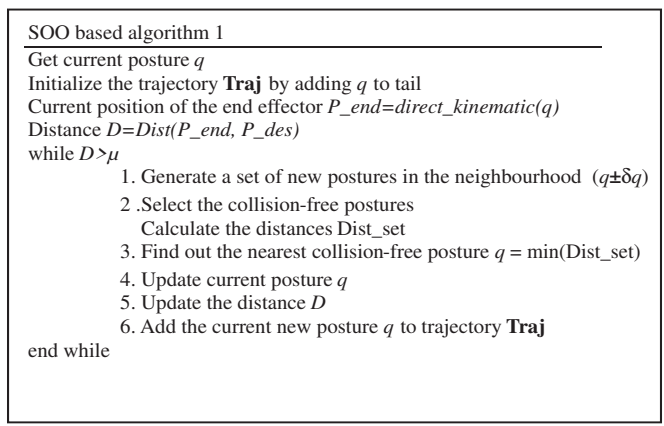}
  \caption{Single objective optimization based trajectory planning method}
  \label{fig:01}       
\end{figure*}

This single objective optimization (SOO) based method was proposed by \cite{RM} in order to generate the trajectory automatically, and its principle is shown in Fig.~\ref{fig:01}. In this algorithm, the distance between the end effector and the destination is chosen as an objective function. For a virtual human, the position of the destination is known and the current posture $q$ is also known. A change ($\pm \delta q$) to the current posture configuration is added to obtain several posture candidates for the next movement. Candidates without collision to the virtual environment are selected out via collision test. The one among the remaining candidates with the smallest distance is selected to update the current posture.
\subsubsection{Limitations of the algorithm}
\label{Sec:2_1_2}
\begin{figure*}
\centering
  \includegraphics[scale=1]{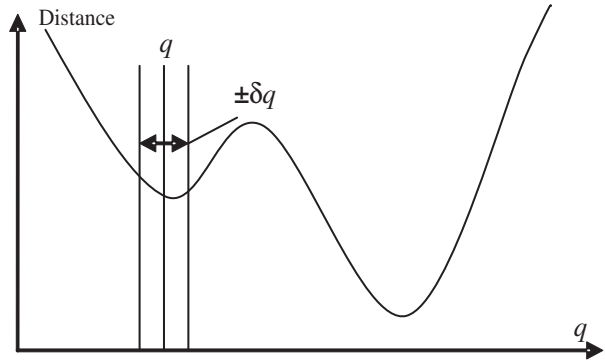}
  \caption{Local minimum in 2-dimension space encountered in single objective optimization based method}
  \label{fig:02}       
\end{figure*}

\begin{figure*}
\centering
  \includegraphics[width=0.75\textwidth]{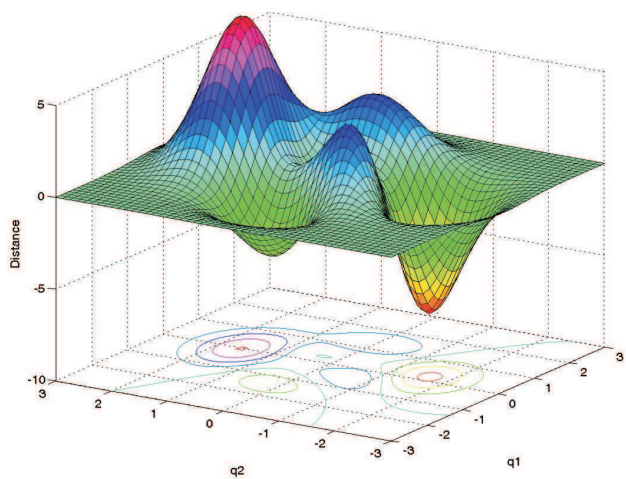}
  \caption{Local minimum encountered in 3-dimension space in single objective optimization based method}
  \label{fig:02_bis}       
\end{figure*}

One of the greatest technical problems in this method is the local minimum encountered while searching the direction to the destination. This problem is illustrated by a simple example in Fig.~\ref{fig:02}. During the trajectory planning, it is very possible that the optimization process will encounter the local minimum. Consequently, this algorithm will be trapped and cannot advance anymore to its global minimum. In this case, the step length $\delta q$ can be modified to skip the local minimum, or the configuration $q$ can be changed by another posture configuration. These modifications need to be done using external intervention \cite{MA2}.
In the situation of Scara Robot, its working space is in 2-dimension space. This problem is illustrated by a simple example in Fig.\ref{fig:02}. In the situation of Staubli Robot, its working space is in 3-dimension space. The problem could be illustrated in Fig.\ref{fig:02_bis}. During the trajectory planning, it is very possible that the optimization process will encounter the local minimum. In this case, this algorithm will be trapped and cannot advance anymore to its global minimum. In this case, the step length $\delta q$ can be modified to skip the local minimum, or the configuration $q$ can be changed by another posture configuration. These modifications need to be done using external intervention.
\subsection{Trajectory planning via external intervention}
\label{Sec:2_2}
\subsubsection{Modified algorithm}
\label{Sec:2_2_1}
\begin{figure*}[!t]
\centering
  \includegraphics[scale=1]{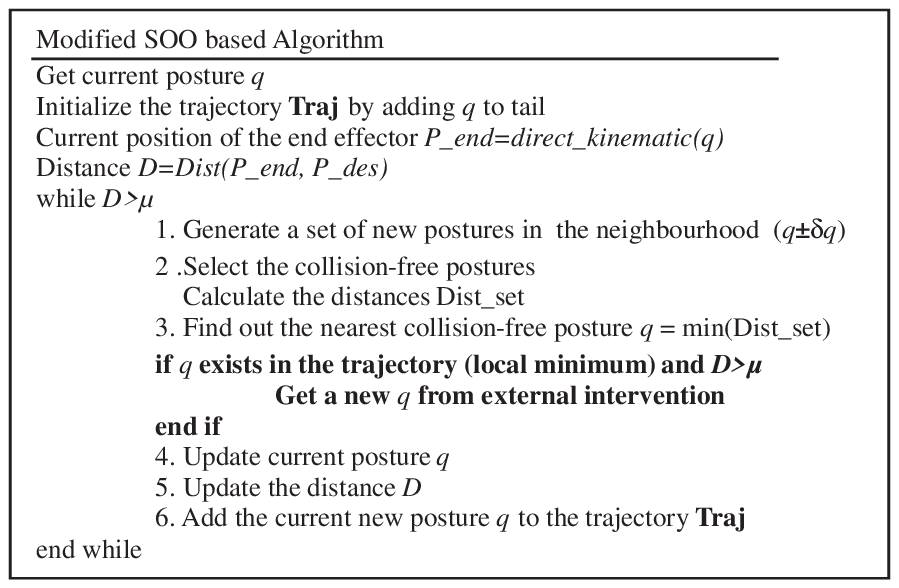}
  \caption{Modified SOO based method}
  \label{fig:03}       
\end{figure*}

As what has been discussed in the section \ref{Sec:2_1}, the single objective optimization method cannot avoid local minimum and that results in no evolution for finding a trajectory to the destination. Therefore, a modified algorithm is proposed in this section using external intervention to overcome this difficulty. The algorithm is presented in Fig.~\ref{fig:03}. Since the step length is constant without intervention, if a posture $q$ has appeared again in the trajectory, there comes local minimum in the trajectory.

\subsubsection{Limitations of the algorithm}
\label{Sec:2_2_2}
\begin{figure*}[!t]
\centering
  \includegraphics[width=0.75\textwidth]{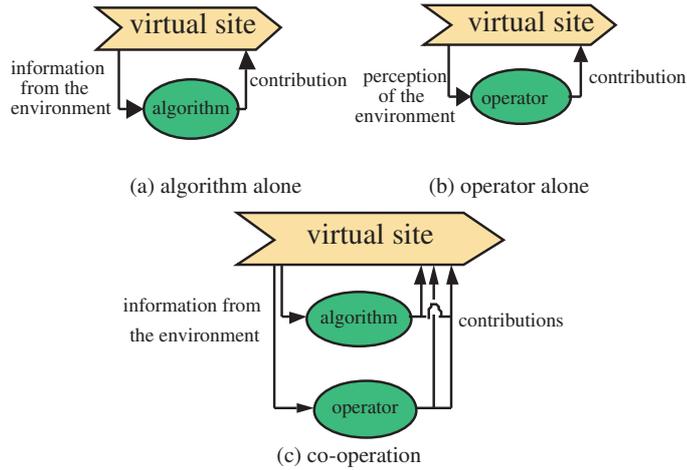}
  \caption{Co-operation principles in multi-agent trajectory planning}
  \label{fig:04}       
\end{figure*}

An external intervention is implemented via different methods. In this part, a multi-agent thought is introduced into our system. The thought is explained by Fig.~\ref{fig:04}. 

In a virtual space, just using algorithm along could plan a path for the virtual human (Fig.~\ref{fig:04}(a)).  At the same time, operating the virtual human directly by changing its rotational configuration manually could also avoid the obstacles to get to the target position (Fig.~\ref{fig:04}(b)). Each way has its disadvantages: using Fig.~\ref{fig:04}(a) will encounter the local minimal obviously; using Fig.~\ref{fig:04}(b) is difficult to generate human-like action and much more time consuming. For these reasons, the co-operator principle (Fig.~\ref{fig:04}(c)) is implemented by us to combine the advantages of each way to get a better result \cite{CC}.
In order to solve these problems, the direct method is to provide a graphical interface to change the configuration: the posture $q$ or the step length $\delta q$. Through changing these parameters the mannequin can be lead to go out of the local minimum situation. Another solution is using motion capture method to skip the local minimum.

In the first solution, changing the posture $q$ can guide the mannequin out of the local minimum situation, but it should be known that it is difficult to have an intuitive manipulation to operate the virtual human by changing the rotational configuration. It requires much more time to manipulate the angular variables directly. Enlarging the step length $\delta q$ can also guide the mannequin to move out the local minimum situation, but the path will not be so smooth.

In the second solution from a motion tracking system, different motion data of a human body can be obtained. The virtual human could be operated much more naturally to skip the local minimum. In the second method, a motion tracking system is required. Accompanying with this method, inverse kinematics or motion retargeting techniques have to be developed to map the motion data to the simulated trajectory.

The main problem is now to define the algorithm which is able to use the principle of the multi-agent system \cite{CC} and to add the information of the motion capture system.
\subsection{Multi-objective evaluation}
\label{Sec:2_3}
\subsubsection{Ergonomic objectives for evaluating the constrained environment}
\label{Sec:2_3_1}
As we have mentioned in section~\ref{Sec:1}, there are different aspects that the designers have to respect. To produce a well designed constrained environment, visibility and accessibility are both important factors. Besides them, the physical influence from the environment should also be assessed in some cases. Therefore, a multi-criteria evaluation system is proposed in this section to evaluate the constrained environment.

{\it Accessibility}: this term describes whether the user could obtain an access to a certain component in the environment. It could be evaluated by the number of possible trajectory solutions (N). The larger the number of solutions, the easier the component can be accessed. If there is no solution for the trajectory, the component is not accessible by a human being.

{\it Visibility}: this term describes the visual accessibility of a component. This term has been modeled or used to analyze workspace in the literature \cite{CC,MJ,SM}. In our research, the visibility is going to be integrated into trajectory planning by treating it as one of the end effectors, since the visible region is also an important factor determining the feasibility of the operation.

{\it Posture effect}: this term describes the effect resulting from posture. This is a traditional subject in ergonomic analysis, and there are several conventional methods for evaluating the posture \cite{ME}. During the ergonomic application, duration of the task, posture engaged in the task, and its physical exposures are taken to evaluate the potential risks of the posture.

{\it Fatigue}: this term is used to describe the effect of physical load on the human body. It has been modeled in \cite{MC2} according to physical exposures related to the manual handling operations. This term is used in our research to evaluate the physical effects of the task realized in the constrained environment. 
\subsubsection{Fatigue evaluation}
\label{Sec:2_3_2}
A muscle model and a recovery model were introduced in \cite{MC2}. These models characterize the capability of a human joint thanks to the knowledge of its history. The fatigue model is defined the following differential Equation
\begin{equation}
			\frac{dF_{cem}(t)}{dt} = -k \frac{F_{cem}(t)}{MVC}F_{load}(t) \label{eq:fatigue}
\end{equation}
\begin{equation}
			\frac{dF_{cem}(t)}{dt} = R(F_{max} - F_{cem}(t)) \label{eq:recovery}
\end{equation}
Related parameters and their descriptions are given in Table \ref{tab:Parameters}. These models describe the muscle fatigue mechanism from a macro aspect based on the muscle motor unit recruitment principle.

\begin{table*}[htp]
	\centering
	\caption{Parameters in muscle model and a recovery model }
	\label{tab:Parameters}
		\begin{tabular}{lcp{0.7\textwidth}}
		\hline
		Item & Unit & Description\\
		\hline
		$MVC$					& $N$ &	Maximum voluntary contraction, maximum capacity of muscle, $F_{max}$\\
		$F_{cem}(t)$ 	& $N$ & Current exertable maximum force, current capacity of muscle\\
		$F_{load}(t)$	& $N$ & External load of muscle, the force which the muscle needs to generate\\
		$k$						& $min^{-1}$ & Constant value, fatigue ratio, here $k=1$\\
		$\%MVC$				&				&Percentage of the voluntary maximum contraction\\
		$f_{MVC}$			&				&$\%MVC/100$, $f_{MVC}=\dfrac{F_{load}}{MVC}$.\\
		$R$           &       &Recovery rate from fatigue model motor units, $R=2.4$ \cite{LIU}\\
		\hline			
		\end{tabular}
\end{table*}

The capacity of each human joint can be described by Eqns.~\ref{eq:fatigue} and \ref{eq:recovery}. For the evaluation of trajectory, a new index is introduce to aggregate the status of each human joint relative to their endurance limit. The remaining forces are defined as, in percentage
\begin{equation}
  F_R= \min\left({\dfrac{F_{{cem}_i}(t)-F_{{Load}_i}}{F_{{MVC}_i}-F_{{Load}_i}}} \right) . 100
\end{equation}
where $F_{{cem}_i}$, $F_{{Load}_i}$ and $F_{{MVC}_i}$ defined the capacity associated with the human joint $i$. These capacities can be divided into two subgroups for the same joint to define the push/pull capacity.
\subsubsection{Technical approach}
\label{Sec:2_3_3}
As discussed before, multi-criteria evaluation system is going to be established to assess different ergonomic aspects of a constrained environment. In this approach, different aspects are mathematically modeled to create objective evaluation. Those results could be useful to improve the design of constrained environment.

\begin{figure*}
\centering
  \includegraphics[width=0.75\textwidth]{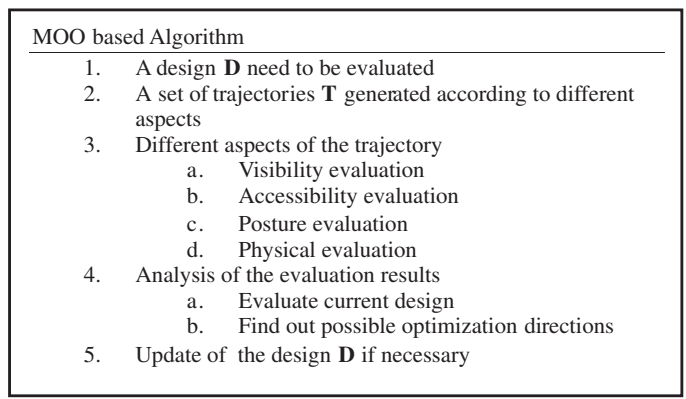}
  \caption{Multi-objective optimization (MOO) based evaluation and trajectory planning algorithm}
  \label{fig:05}       
\end{figure*}

Meanwhile, a multi-objective optimization procedure could also be interesting to determine design parameters of a constrained environment. This algorithm is presented in Fig.~\ref{fig:05}. 

\section{Case study}
\label{Sec:3}
\subsection{Hardware and software implementation}
\label{Sec:3_1}
In order to realize our algorithm, a virtual reality platform is constructed. This platform includes two parts: simulation system and motion capture system. The simulation system is mainly responsible for the generation of virtual environment, the collision computation, and the automatic trajectory planning. An optical motion tracking system is in charge of capturing motion data and communicating with the simulation system.
 
\begin{figure*}
\centering
  \includegraphics[width=0.75\textwidth]{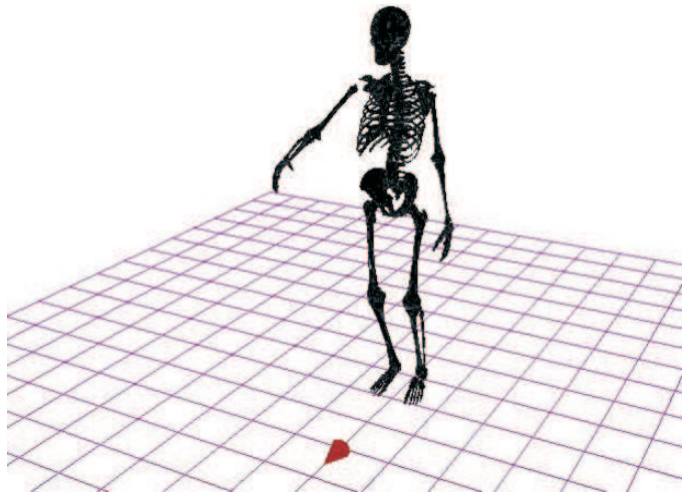}
  \caption{The virtual skeleton in the simulation system}
  \label{fig:06}       
\end{figure*}
The simulation system is developed using OpenGL and C++. The virtual skeleton is shown graphically in Fig.~\ref{fig:06}. The virtual human is combined by ten body parts: head, torso, thighs, shanks, upper arms, and lower arms. Each body part is modeled as a 3DS model file which is composed of hundreds of triangle facts. The virtual skeleton is driven using direct kinematic method by changing angular values of each key joints. Meanwhile, virtual environments could also be loaded from 3DS files which are converted directly from CAD models.

\begin{figure*}
\centering
  \includegraphics[width=0.75\textwidth]{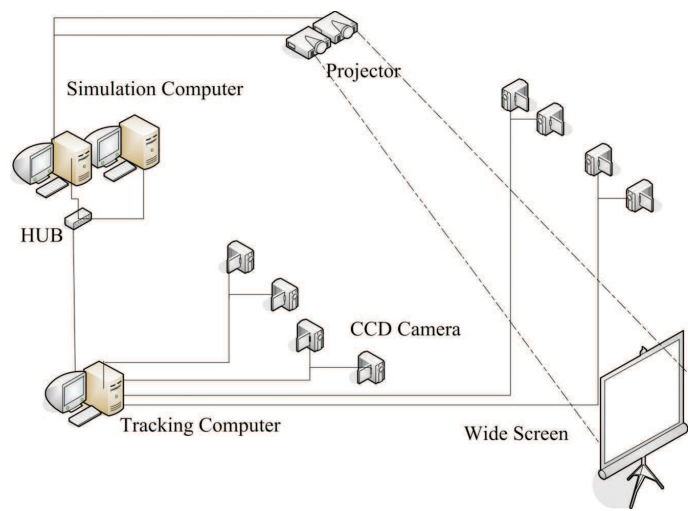}
  \caption{Optical motion capture system}
  \label{fig:07}       
\end{figure*}

In the motion capture system (Fig.~\ref{fig:07}), there are totally eight CCD cameras to capture the motion in a range of $2m \times 2m \times  2m$. Nonlinear direct transformation method is used to calibrate all the cameras. After calibration, the system could capture maximum 13 markers at 25Hz \cite{MZ2}. Although this tracking speed is not enough for capturing accurate motion, it still provides an acceptable speed to adjust the posture.

\subsection{Robot trajectory planning for Scara Robot}
\label{Sec:3_2}
\begin{figure*}
\centering
  \includegraphics[width=0.75\textwidth]{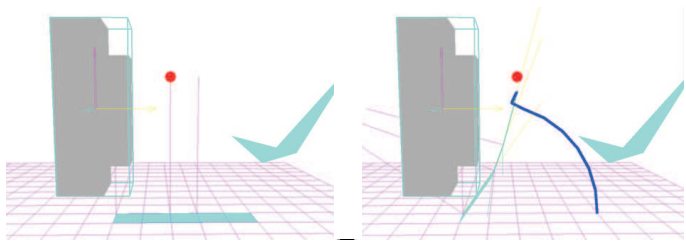}
  \caption{Trajectory planning test using a Scara robot (with link length equals 20, 10, 20)}
  \label{fig:08}       
\end{figure*}

\begin{figure*}
\centering
  \includegraphics[width=0.75\textwidth]{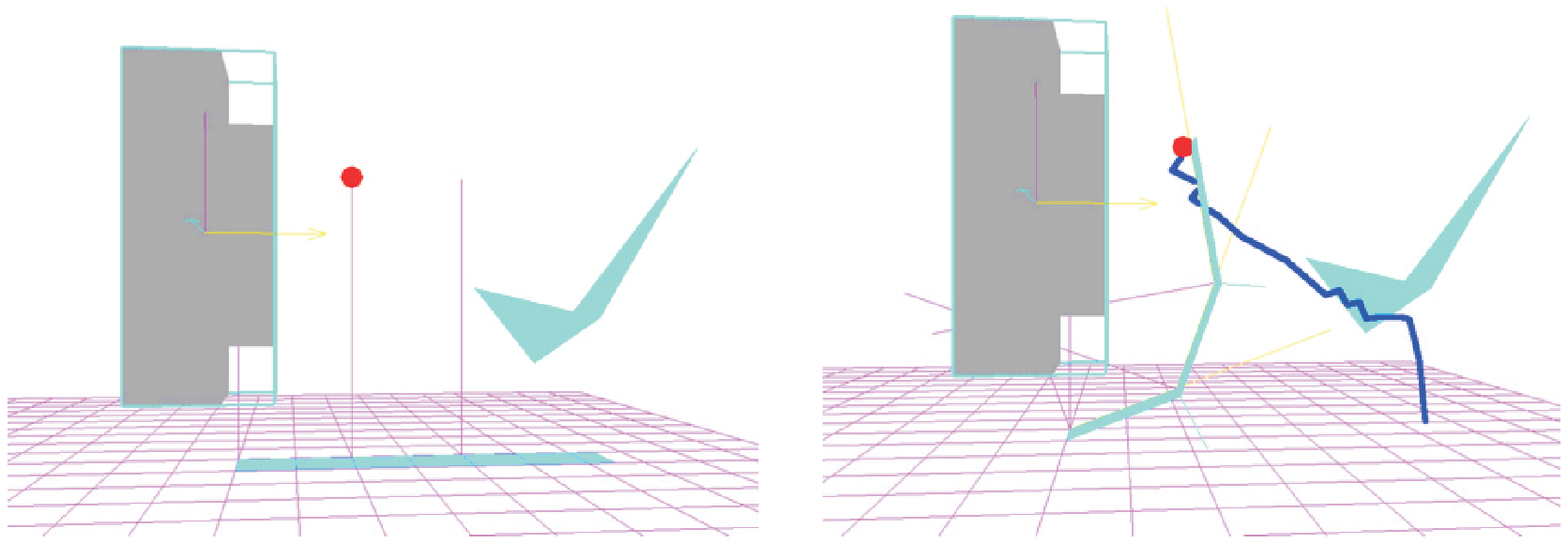}
  \caption{Trajectory planning test using a Scara robot (with link length equals 20, 20, 25)}
  \label{fig:09}       
\end{figure*}

\begin{figure*}
\centering
  \includegraphics[width=0.75\textwidth]{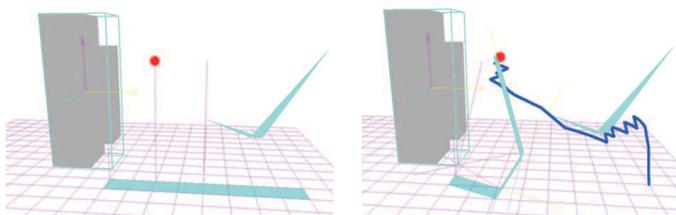}
  \caption{Trajectory planning test using a Scara robot (with link length equals 20, 20, 40)}
  \label{fig:10}       
\end{figure*}

Since there are too many degrees of mobilities in a virtual human, at the very beginning of our research, a trial demonstration of the algorithm has been realized by using a Scara robot and several virtual objects in a virtual environment (see Fig.~\ref{fig:08}--\ref{fig:10}). The robot is composed of three rectangles, and the end effector is the right end. The blue round point denotes the destination of the end effector. There are three obstacles (two triangles and one block) in the environment. 

Different trajectories will be generated by using our algorithm according to the different length of each revolute joint in the Scara robot. Figure~\ref{fig:08} shows that trajectory of Scara robot with the link length parameters (20, 10, 20). Figure~\ref{fig:09} shows that the trajectory of Scara robot with the link length parameters (20, 20, 25) and Fig 10 shows that the trajectory of Scara robot with the link length parameters (20, 20, 40). From these figures, we can see that in the same environment, different size of the Scara robot can come across various situations. This is necessary and also important, because of a product is not just for a fixed user. Various situation or parameters of a subject should be taken into consideration. The lengths of the links are changed demonstrate different effects of dimensional information.

The trajectory in blue is generated by the algorithm presented in section \ref{Sec:2_1_1}. It is observable that the trajectory in blue could avoid the collision while approaching to the destination. Figures~\ref{fig:08}, \ref{fig:09} and \ref{fig:10} show that with the same obstacle different geometrical configurations (different arm lengths) can generate different paths to avoid the obstacle from the same start to the same destination. 

\begin{figure}[hbt]
    \begin{center}
    \begin{tabular}{cc}
       \begin{minipage}[t]{55 mm}
         \includegraphics[width=0.9\textwidth]{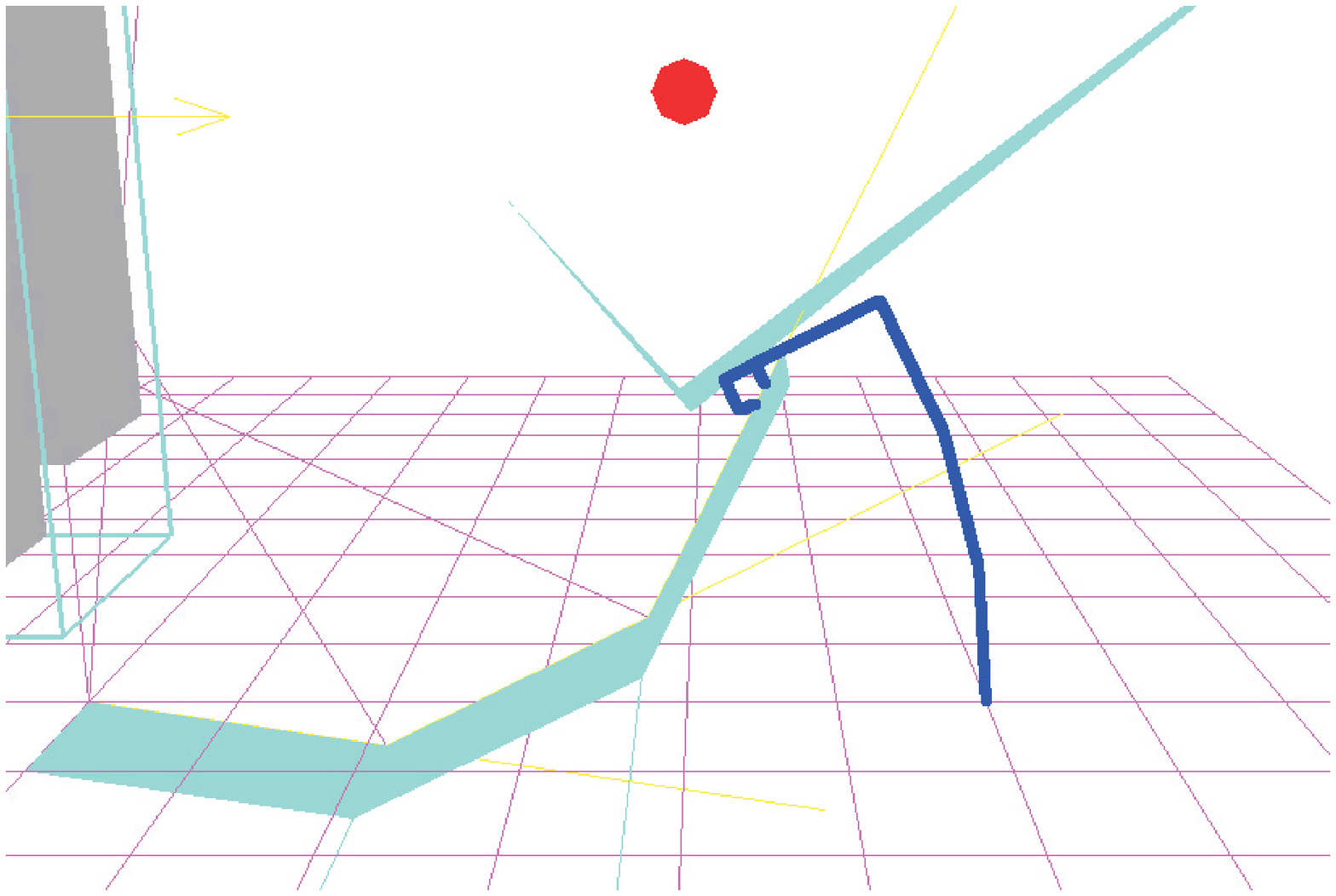}
         \caption{Trajectory planning test using a Scara robot (with local minimum, step=0.08)}
          \label{fig:11}       
      \end{minipage} &
       \begin{minipage}[t]{55 mm}
         \includegraphics[width=0.9\textwidth]{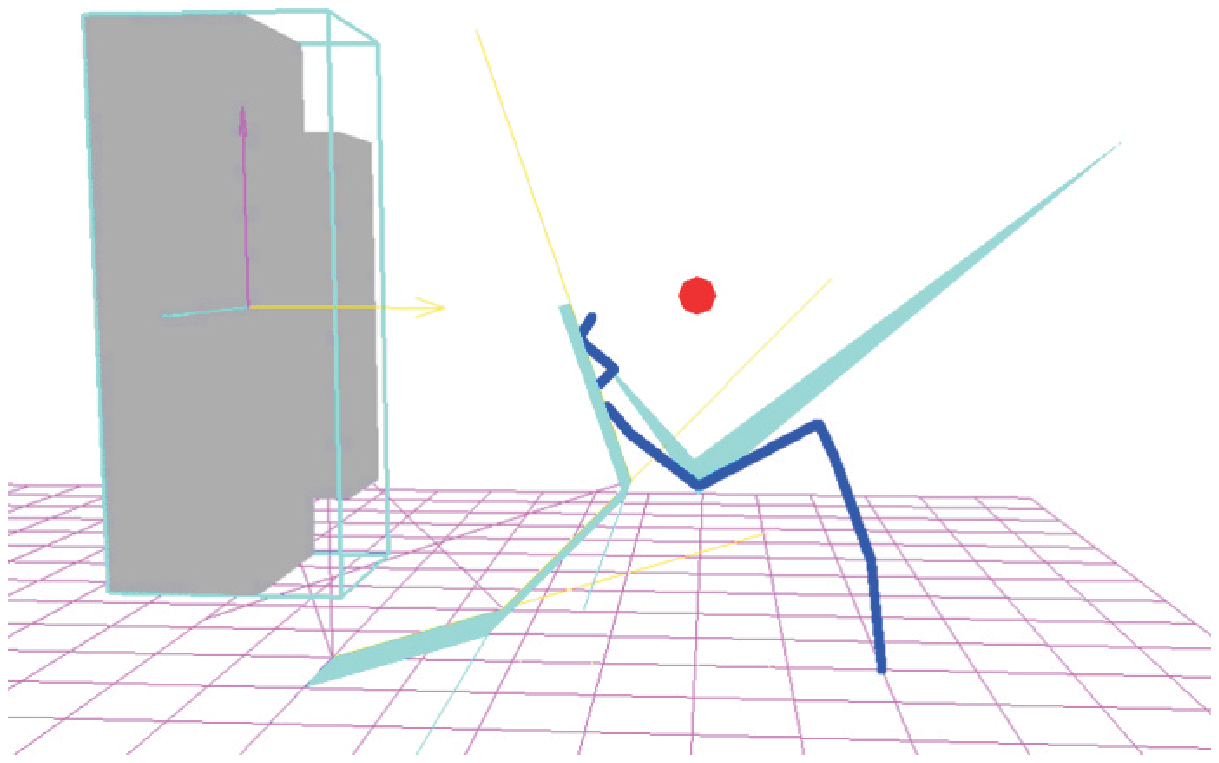}
         \caption{Trajectory planning test using a Scara robot (External intervention interface, step=0.1)}
         \label{fig:12}       
      \end{minipage}
    \end{tabular}
    \end{center}
\end{figure}

In the obstacle avoiding process, there is always possibility of local minimum. In Fig.~\ref{fig:11}, a demonstration of local minimum is shown. 

Since the obstacles locate between the destination and the end effector and the descending direction is also restricted by the obstacles, the algorithm could not skip the local minimum with a step length 0.08.

In Fig.~\ref{fig:12}, for the same arrangement of obstacles in Fig 10, the step length has been adjusted to 0.1. As a result, the first obstacle could be passed over without problem.
\subsection{Robot trajectory planning for Staubli Robot}
\begin{figure*}
\centering
  \includegraphics[width=0.75\textwidth]{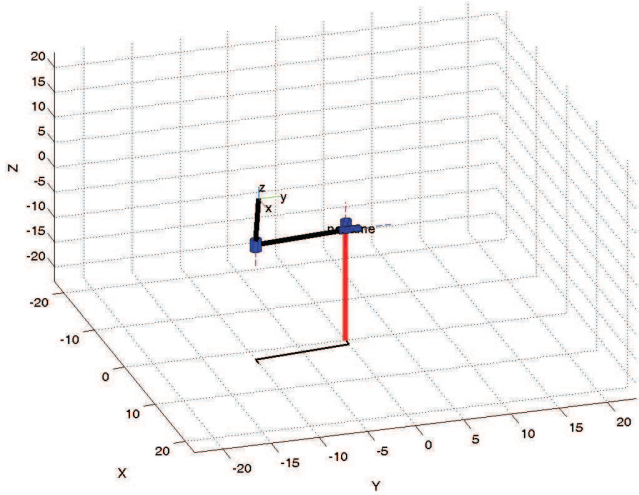}
  \caption{Geometry model of arm}
  \label{fig:13}       
\end{figure*}

In the trial demonstration of the Scara Robot, its working space is in 2-dimensional. For each articulation it just has one degree of freedom. In this situation, the local minimum problem is usually happened. The second trial of the algorithm is in the Staubli robot, we use the part of shoulder and elbow,  in shoulder it has 2 degrees of freedom and in the elbow it has 1 degree of freedom. We do this trial demonstration because of this robot is more similar to the human arm. The Arm geometric model is in Fig.~\ref{fig:13}. The geometry modelling parameter of arm is in Fig.~\ref{fig:13}. $d_3$ is the length of upper arm, $d_4$ is the length of forearm. $\theta_1, \theta_2$ is the rotation angle of shoulder in horizontal and vertical direction. $\theta_3$ is the rotation angle of elbow.

\begin{figure*}
	\centering
  \includegraphics[width=0.75\textwidth]{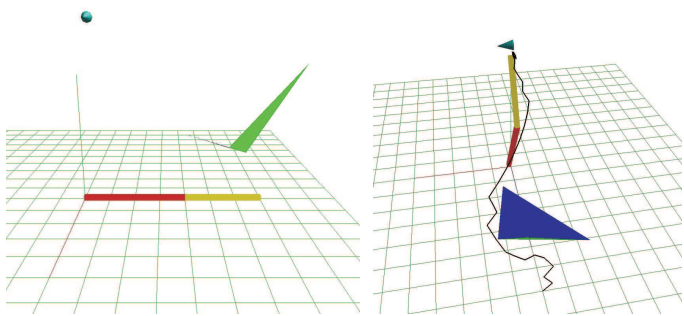}
  \caption{Trajectory planning test using a Staubli robot (with link length equals 40,30)}
  \label{fig:14}       
\end{figure*}

\begin{figure*}
\centering
  \includegraphics[width=0.75\textwidth]{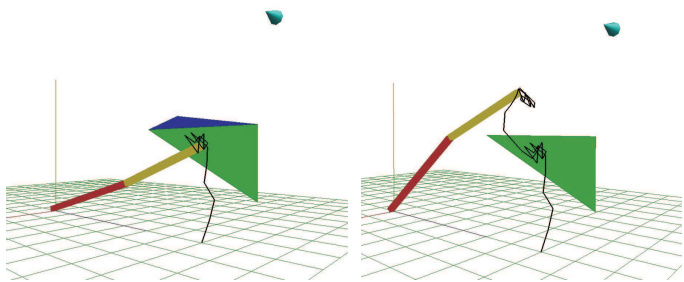}
  \caption{Local minimum problem of Staubli robot}
  \label{fig:15}       
\end{figure*}

\begin{table}
  \centering
  \caption{Geometry modeling parameters of arm}
  \label{tab:geo}
  \begin{tabular}{|c|c|c|c|c|c|}
  \hline
  $j$ & $\sigma_j$ & $\alpha_j$&$d_j$&$\theta_j$&$r_j$\\
  \hline
  1&0&$0$&0&$\theta_1$&0\\
  2&0&$\pi/2$&0&$\theta_2$&0 \\
  3&0&0&$d_3$&$\theta3$&0\\
  4&0&0&$d_4$&0&0\\
  \hline
  \end{tabular}
\end{table}
Based on the Staubli robot, 3-dimensional trajectorie will be   generated by using our algorithm. Figure~\ref{fig:14}. shows that trajectory of  Staubli robot with the link length parameters (40, 30) when it want to reach the destination how to avoid a spatial obstacle and the trajectory is in 3-dimenstional. The black line is the trajectory. Because of the Staubli robot in the shoulder has 2 degree of freedom, when it meet the obstacle, it has one more dimension direction to choose, this means that it has more chance to avoid the obstacle compared with the Scara Robot. Although the 3-dimensional direction could improve the solution when the arm meets the obstacle, it also has the possibility to be locked in the local minimum problem. The Fig.~\ref{fig:14} shows that the obstacle is composed by 2 triangles, and it forms a concave shape. When the arm moves inside the concave, it is locked. From the Fig.~\ref{fig:15} we could see that it repeat the irregular trajectory inside the concave. After change the step=0.06, it jumps out the local minimum range and move to the destination.
    
\subsection{Virtual human trajectory planning}
\label{Sec:3_3}
Virtual human trajectory planning using the proposed algorithm is still under construction. There are still several steps to complete the demonstration: motion retargeting, modeling of different aspects, and the complete installation of the motion capture system.

\begin{figure*}
  \includegraphics[width=0.75\textwidth]{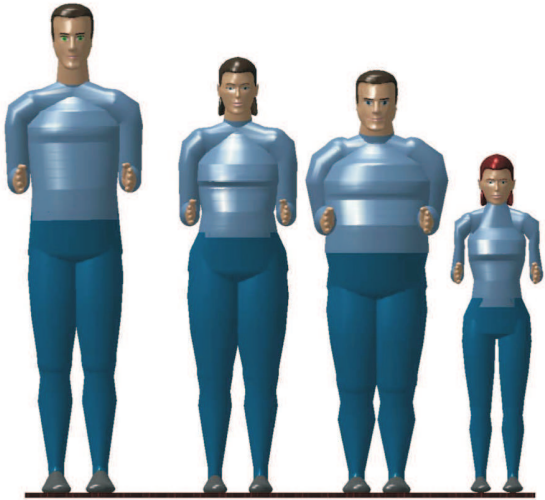}
  \caption{A set of mannequins with different percentiles to test new products}
  \label{fig:16}       
\end{figure*}

The definition of the trajectories should be not just used for one person, because of a product is designed for a given population (Fig.~\ref{fig:16}). An automatic path planner can calculate the path from a start point to a destination. But imaging that in a complex environment and for many users, sometimes the algorithm might be failed. In this situation, the user interaction has to be limited to minimize design effort. 
  
\begin{figure*}
\centering
  \includegraphics[width=0.75\textwidth]{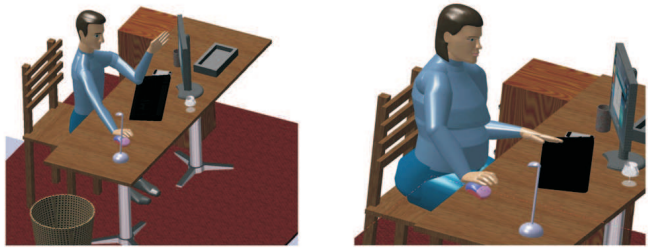}
  \caption{Example of office design for two different female cases}
  \label{fig:17}       
\end{figure*}

The ergonomic study of the product can be tested thanks to the definition of a set of tasks: taking a mouse, touching the screen (Fig.~\ref{fig:17}). Different tasks have different aspects to be respected, and different aspects will result different trajectories.
Different sizes, different weights, different tasks, and different design aspects: all these factors lead trajectory planning for a virtual human to a quite difficult problem. The multi-objective evaluation and optimization approach for virtual human trajectory planning has to be developed with caution. 
\section{Conclusions and perspectives}
In this paper, an approach of using virtual human in constrained environment design was presented, and a new algorithm was proposed to solve trajectory planning for two types of robot. Our preliminary application of the algorithm demonstrates that this interactive method had a potential to help us solve trajectory planning problem in constrained environment for mannequins. In our future research, a multi-agent system and multi-objective optimization method will be implemented to facilitate the design process of constrained environments.
\begin{acknowledgements}
This research is supported in the context of collaboration between the Ecole Centrale de Nantes (Nantes, France) and Tsinghua University (Beijing, P.R.China). The lead author would also like to thank \'Ecole Centrale de Nantes for the financial support of the post-doctorate studies. The authors would like to address thanks to the financial support of \'Ecole Centrale de Nantes, R\'egion des Pays de la Loire in the project Virtual Reality for design (VR4D) and China Scholarship Council (CSC).
\end{acknowledgements}



\end{document}